\documentclass[letterpaper]{article} 
\usepackage{aaai24}  
\usepackage{times}  
\usepackage{helvet}  
\usepackage{courier}  
\usepackage[hyphens]{url}  
\usepackage{graphicx} 
\usepackage{natbib}  
\usepackage{caption} 
\usepackage{algorithm}
\usepackage{algorithmic}

\usepackage{svg}
\usepackage{multirow}

\usepackage{tabularx}
\usepackage{booktabs}
\usepackage{xcolor}
\usepackage{pifont}
\usepackage{bbding}
\usepackage{amssymb}
\usepackage{amsfonts}
\usepackage{mathrsfs}
\usepackage{newfloat}
\usepackage{listings}
\DeclareCaptionStyle{ruled}{labelfont=normalfont,labelsep=colon,strut=off} 
\floatstyle{ruled}
\newfloat{listing}{tb}{lst}{}
\floatname{listing}{Listing}
\title{ResDiff: Combining CNN and Diffusion Model for Image Super-Resolution}
\author {
    Shuyao Shang \textsuperscript{\rm 1},
    Zhengyang Shan \textsuperscript{\rm 1},
    Guangxing Liu \textsuperscript{\rm 1},
    LunQian Wang \textsuperscript{\rm 2},
    XingHua Wang \textsuperscript{\rm 2}, \\
    Zekai Zhang \textsuperscript{\rm 3},
    Jinglin Zhang$^*$ \textsuperscript{\rm 1}
}
\affiliations {
    \textsuperscript{\rm 1} Shandong University\\
    \textsuperscript{\rm 2} Linyi University\\
    \textsuperscript{\rm 3} Qilu University of Technology\\
    202000800098@mail.sdu.edu.cn, 
    202000800128@mail.sdu.edu.cn, 
    202000800141@mail.sdu.edu.cn,
    210854002032@lyu.edu.cn,
    210854002052@lyu.edu.cn,
    10431210745@stu.qlu.edu.cn,
    jinglin.zhang@sdu.edu.cn
}

\usepackage{bibentry}

\begin{document}

\maketitle

\begin{abstract}
Adapting the Diffusion Probabilistic Model (DPM) for direct image super-resolution is wasteful, given that a simple Convolutional Neural Network (CNN) can recover the main low-frequency content. Therefore, we present \textbf{ResDiff}, a novel \textbf{Diff}usion Probabilistic Model based on \textbf{Res}idual structure for Single Image Super-Resolution (SISR). ResDiff utilizes a combination of a CNN, which restores primary low-frequency components, and a DPM, which predicts the residual between the ground-truth image and the CNN-predicted image. In contrast to the common diffusion-based methods that directly use LR space to guide the noise towards HR space, ResDiff utilizes the CNN's initial prediction to direct the noise towards the residual space between HR space and CNN-predicted space, which not only accelerates the generation process but also acquires superior sample quality. Additionally, a frequency-domain-based loss function for CNN is introduced to facilitate its restoration, and a frequency-domain guided diffusion is designed for DPM on behalf of predicting high-frequency details. The extensive experiments on multiple benchmark datasets demonstrate that ResDiff outperforms previous diffusion-based methods in terms of shorter model convergence time, superior generation quality, and more diverse samples.
\end{abstract}

\section{Introduction}

Single Image Super-Resolution (SISR) is a difficult task in computer vision, which aims to recover high-resolution (HR) images from their low-resolution (LR) counterparts. During image degradation, the high-frequency components are lost, and multiple HR images could produce the same LR image, making this task ill-posed. After Generative Adversarial Networks(GAN) \cite{gan} was proposed, the main generative-model-based SISR methods are GAN-driven. However, GAN-based methods are hard to train and prone to fall into pattern collapse, causing a lack of diversity. Therefore, a superior generative model is required in the SISR task.

\begin{figure*}[htbp]
\begin{center}
\includegraphics[width=16 cm]{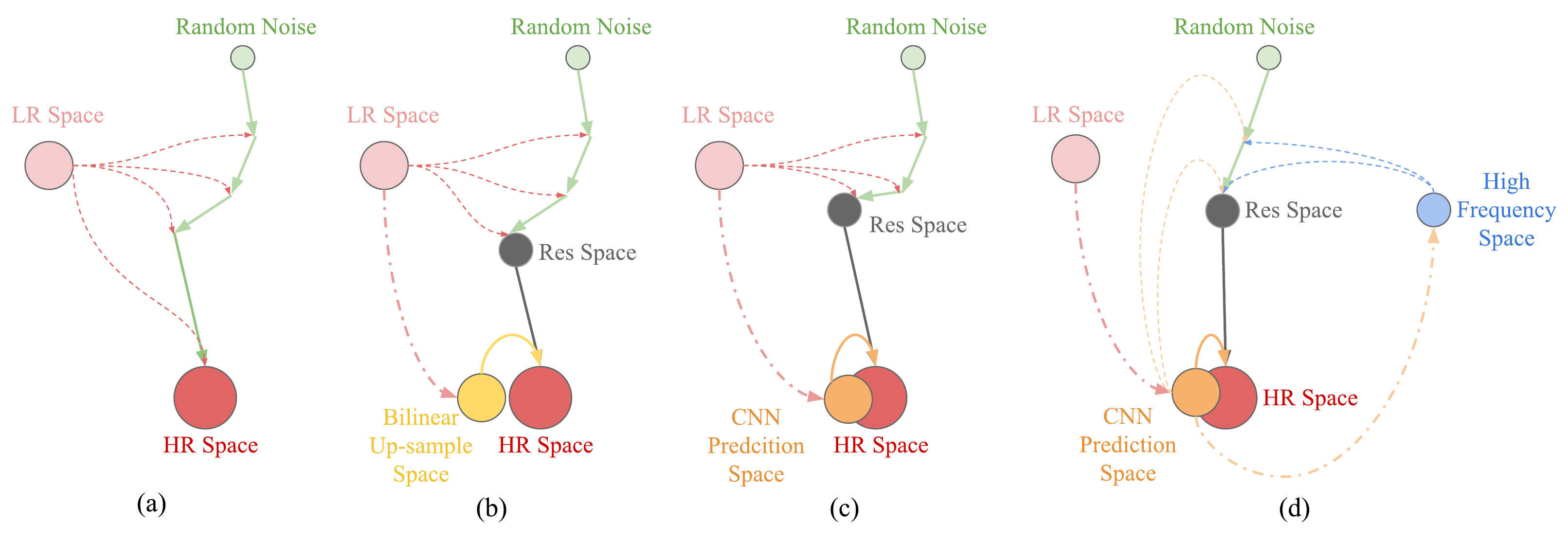}
\end{center}
\caption{Comparison of different generation processes. In contrast to (a) \cite{sr3}, (b) \cite{srdiff}, (c) \cite{deblur} where only LR Space is used to guide the generation, our ResDiff (d) makes full utilization of CNN Prediction Space and High-Frequency Space to guide a faster and better generation.}
\label{fig:intro_compare}
\end{figure*}

Diffusion Probabilistic Model (DPM) has already demonstrated impressive capabilities in image synthesis \cite{imagen, dpm_gen, latent-diff, unclip} and image restoration \cite{ILVR,DDRM,DDNM}. It has also shown promising prospects in SISR tasks \cite{sr3,srdiff}. However, current Diffusion-based methods for SISR, such as SR3\cite{sr3}, generate HR images directly from random noise, and LR images are only used as conditional input to the diffusion process (Fig.\ref{fig:intro_compare} 
(a)). Consequently, the diffusion model needs to recover both the high and low-frequency contents of the image, which not only prolongs the convergence time but also inhibits the model from focusing on the fine-grained information, potentially missing texture details. Li et al.\cite{srdiff} had taken this into account but employed only a bilinear interpolation for the initial prediction, which, compared to CNN, failed to restore sufficiently low-frequency contents and was incapable of generating any high-frequency components in the initial prediction (Fig.\ref{fig:intro_compare} (b)). Similarly, whang et al.\cite{deblur} designed a random-sampler and a deterministic-predictor to tackle this problem. However, there is no information interaction between the random-sampler and the deterministic-predictor, resulting in the latter not functioning to its full potential (Fig.\ref{fig:intro_compare} (c)).

\begin{figure}[tbp]
\begin{center}
\includegraphics[width=8 cm]{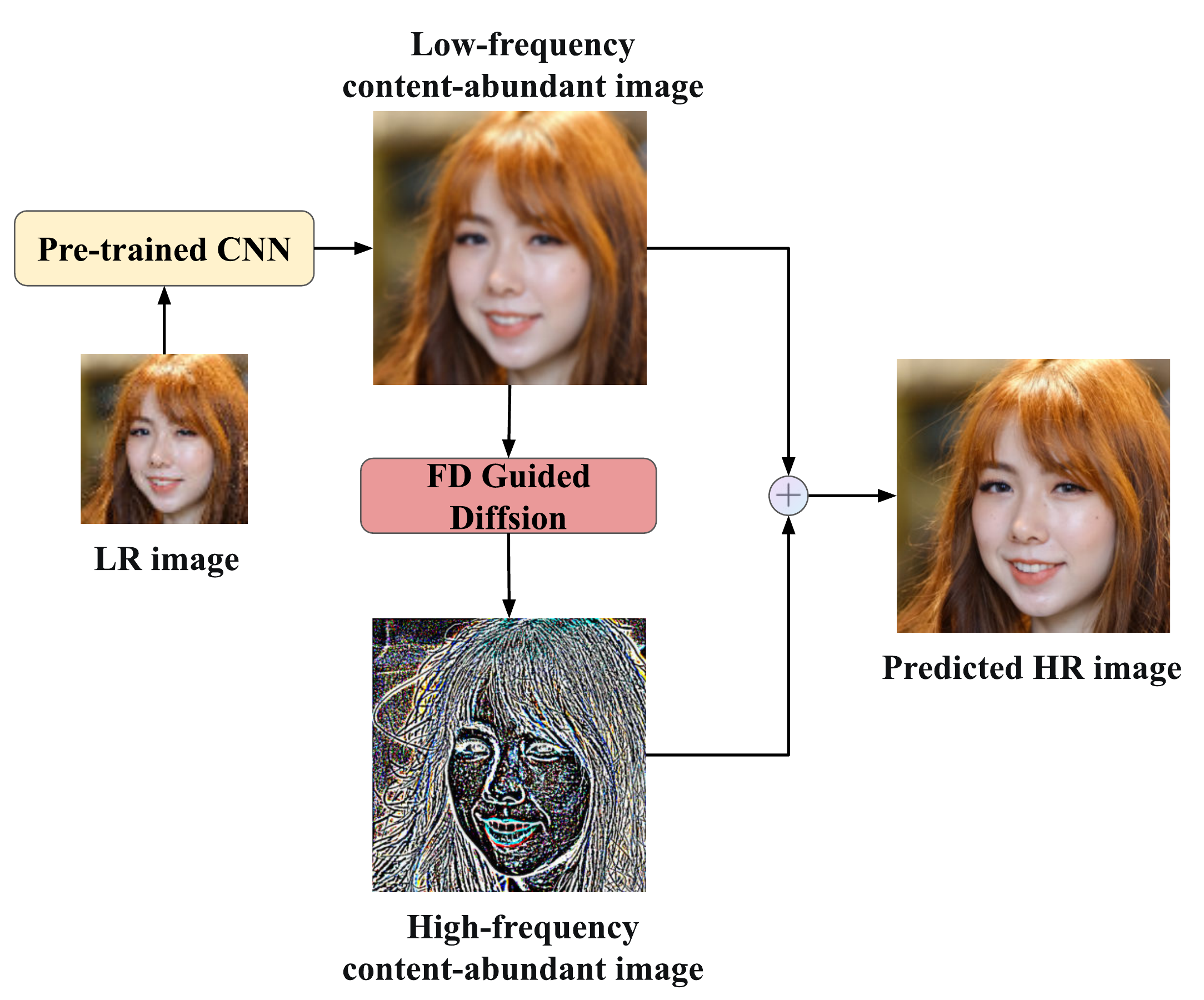}
\end{center}
\caption{Overall struture of proposed ResDiff.}
\label{fig:overall}
\end{figure}

Inspired by the above \cite{srdiff,deblur}, we propose ResDiff, a residual-structure-based diffusion model. Unlike \cite{srdiff}, ResDiff utilizes a CNN for initial prediction. And in contrast to \cite{deblur}, the CNN in ResDiff is pre-trained, thus capable of restoring the major low-frequency components and partial high-frequency components. The initial prediction of the CNN is adopted to guide the random noise towards the Res Space (i.e., the residual space between the Ground Truth image and the CNN predicted image). Compared to the methods that only use LR space as guidance, ResDiff can leverage additional information and generate richer high-frequency details. (Fig.\ref{fig:intro_compare} (d)). Fig.\ref{fig:overall} presents the structure of ResDiff. The CNN used in ResDiff contains a limited number of parameters. Thus, two more loss functions are introduced to strengthen its recovery capabilities. To further enhance the generation quality, we design a Frequency Domain-guided Diffusion (FD-guided Diffusion) as shown in  Fig.\ref{fig:intro_compare} (d) where the high-frequency space also guides the generation process. FD-guided Diffusion consists of two novel modules. The first is a Frequency-Domain Information Splitter (FD Info Splitter) that separates high-frequency and low-frequency contents and performs adaptive denoising on the noisy image. The second is a high-frequency guided cross-attention module (HF-guided CA) that helps the diffusion model predict high-frequency details. The pseudo-code for sampling with ResDiff is as Alg.\ref{alg:2}.

Experiments on two face datasets (FFHQ and CelebA) and two general datasets (Div2k and Urban100) demonstrate that ResDiff not only accelerates the model's convergence speed but also generates more fine-grained images. To verify the generalization of our method, more experiments on different types of datasets \cite{LSCIDMR} are given in the supplementary material.

Our contributions can be summarized as follows:

• \textbf{Shorter Convergence Time}: We have designed ResDiff, a residual structure-based diffusion model for the SISR task that leads to an apparent improvement in convergence speed compared to other diffusion-based methods.

• \textbf{Superior Generation Quality}: We have introduced FD-guided Diffusion to enhance the diffusion model's concentration on high-frequency details, resulting in superior generation quality.

• \textbf{More Diverse Output}: Experiments have demonstrated that ResDiff holds a lower perceptual-based evaluation value, indicating our method is capable of producing diverse samples.


\section{Related Works}

Generative-model-based methods have created great success in SISR, which can be classified into GAN-based\cite{srgan,esrgan,DPSRGAN,SFT-GAN,ranksrgan}, flow-based\cite{srflow,Conditional-Flow}, and fiffusion-based\cite{sr3,srdiff} methods. 


\paragraph{GAN-based methods} Ledig et al.\cite{srgan} proposed SRGAN, which employs a perceptual loss function to generate high-quality images. Similarly, Kim et al.\cite{esrgan} introduced ESRGAN, which adopted an enhanced super-resolution GAN and a superior loss function to improve the perceptual quality. GAN-based methods combine content losses with adversarial losses, allowing them to generate sharp edges and richer textures. However, they are prone to mode-collapse, which decreases diversity in the generated SR samples. Moreover, training GANs is challenging and may lead to unexpected artifacts in the generated image.


\paragraph{Flow-based methods} Lugmayr et al.\cite{srflow} proposed SRFlow, which is a flow-based method that learns the conditional distribution of high-resolution images given their low-resolution counterparts, enabling high-quality image super-resolution with natural and diverse outputs. Flow-based methods map HR images to flow-space latents using an invertible encoder and connect the encoder and decoder with an invertible flow module, which avoids training instability but requires higher training costs and provides lower perceptual quality. 


\paragraph{Diffusion-based methods} Li et al.\cite{srdiff} introduced SrDiff, the first diffusion-based model for SISR, demonstrating that using the diffusion model for SISR tasks is feasible and promising. Saharia et al. proposed Sr3 \cite{sr3}, which adapts Denoising Diffusion Probabilistic Models (DDPM) to perform SISR tasks, yielding a competitive perceptual-based evaluation value. Diffusion-based methods utilize a diffusion process that simulates noise reduction, resulting in sharper and more detailed images. However, a high computational cost is needed due to multiple forward and backward passes through the entire network during the training process. Our proposed ResDiff, though without improving the training speed of a single iteration, accelerates convergence, which can alleviate this issue from another perspective.

\begin{algorithm}[tb]
\caption{ResDiff Inference}
\label{alg:2}
\textbf{Input}: low-resolution image $x_{LR}$ and pre-trained CNN \\
\textbf{Parameter}: $\mu_\theta$ and $\Sigma_\theta$ same as in DDPM\\
\textbf{Output}: High-resolution image
\begin{algorithmic}[1] 
    \STATE $x_{cnn}$ = CNN($x_{LR}$)
    \STATE $x_T \sim \mathcal{N}(0,I)$   
    \STATE $\textbf{for} \ t=T:1 \ \textbf{do}$
    \STATE $ \quad \epsilon \sim \mathcal{N}(0,I) \ $if$ \ t>1, \ $else$ \ \epsilon=0 $
    \STATE $\quad x_{t-1}= \mu_\theta(x_t,t,x_{cnn}) + \sqrt{\Sigma_\theta(x_t,t,x_{cnn})} \ \epsilon  $
    \STATE $\textbf{end} \ \textbf{for}$
    \STATE $\textbf{return} \ x_0+x_{cnn}$
\end{algorithmic}
\end{algorithm}


\section{The Proposed ResDiff}

\subsection{Pre-trained CNN}

\begin{figure}[t]
\begin{center}
\includegraphics[width=8.5 cm]{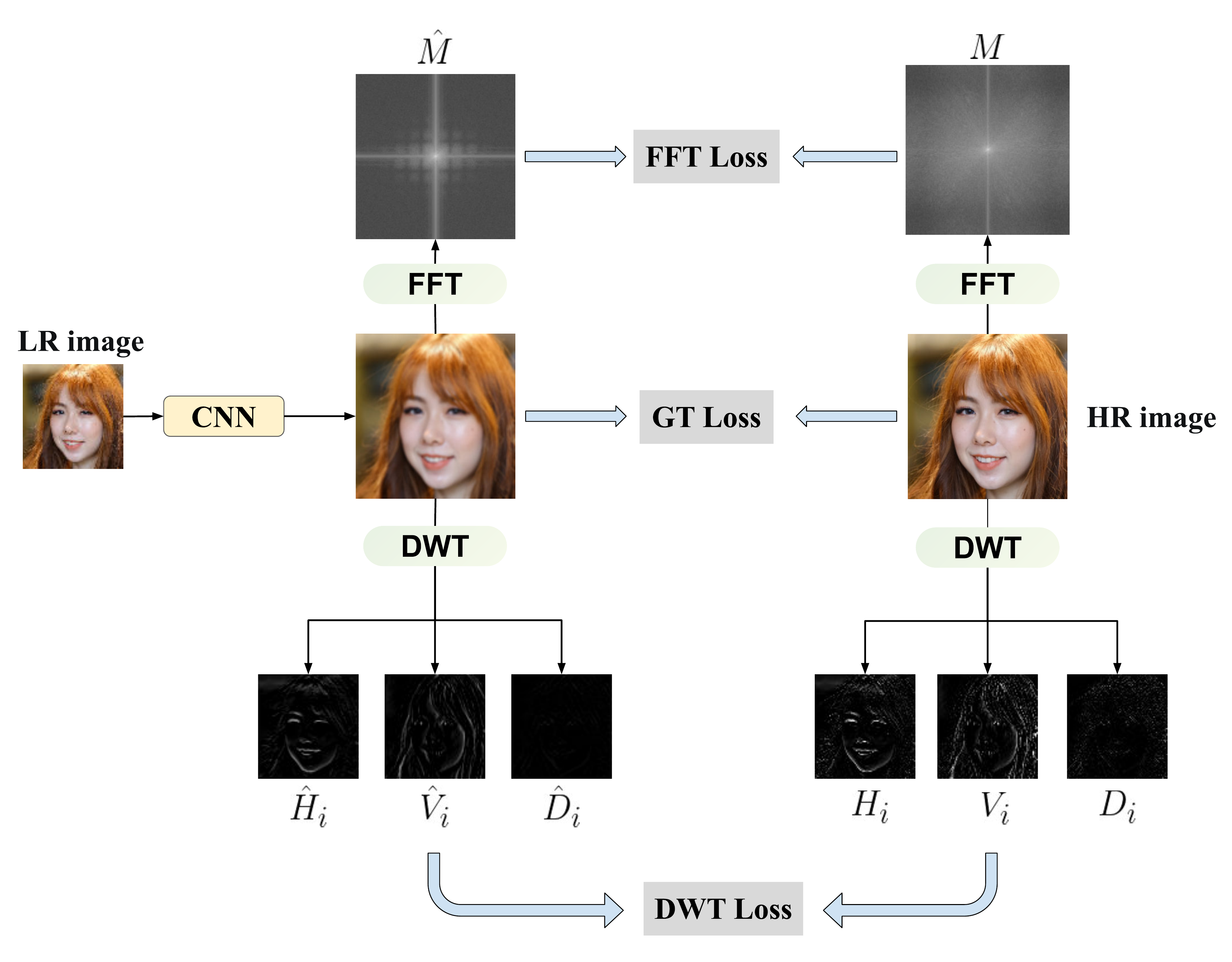}
\end{center}
\caption{Depiction of the three loss functions utilized in CNN pre-training. A spatial domain loss (GT Loss) and two frequency domain losses (FFT Loss and DWT Loss) are computed.}
\label{fig:cnn_loss}
\end{figure}

To reduce additional training costs, we utilize a CNN with a reduced number of parameters to generate an initial prediction. This CNN aims to recover primary low-frequency components and partial high-frequency components, consequently facilitating the diffusion model's restoration of the more intricate high-frequency details. To ensure its generating capability, we are enlightened by \cite{dwt_loss, dwt_loss_2} and introduce two more loss functions (Fig.\ref{fig:cnn_loss}), namely $\mathcal{L}_{FFT}$ based on the Fast Fourier Transform (FFT) \cite{fft} and $\mathcal{L}_{DWT}$ based on the Discrete Wavelet Transform (DWT) \cite{dwt}, in addition to the original loss function.

The $\mathcal{L}_{FFT}$ can be defined as the mean square error(MSE) between the magnitudes of the FFT coefficients of the two images:


\begin{equation}
    \mathcal{L}_{FFT} = \mathbb{E}[\left\| M - \hat{M} \right\|^2]
\end{equation}

where $M$ and $\hat{M}$ denote the frequency domain images obtained by performing FFT on the ground-truth image and the predicted image.

In a bid to enable the CNN to further recover partial high-frequency contents on top of recovering the primary low-frequency contents, we designed $\mathcal{L}_{DWT}$. Performing DWT on an image will decompose it into four sub-bands: low-low (LL), low-high (LH), high-low (HL), and high-high (HH). LL sub-band contains the low-frequency content of the image, while the remaining three contain the high-frequency components of the image from horizontal, vertical, and diagonal directions, respectively. The LL sub-band can perform further similar decomposition to obtain multi-layer high-frequency components. As for $\mathcal{L}_{DWT}$, we extract the wavelet coefficients of the high-frequency bands $H$, $V$, and $D$, which refer to the high-frequency components in the horizontal, vertical, and diagonal directions, respectively. For both the ground-truth image and predicted image, $\mathcal{L}_{DWT}$ compute the MSE between each high-frequency sub-band:


\begin{equation}
    \mathcal{L}_{DWT} = \sum_{i=1}^{L}\mathbb{E} [ \left\|\hat{H_i}-H_i\right\|^2 + \left\|\hat{V_i}-V_i\right\|^2 +\left\|\hat{G_i}-G_i\right\|^2  ]
\end{equation}

where $H_i$,$V_i$,$D_i$ are the sub-bands of the ground-truth image in the $i$-th downsampling, and $\hat{H_i}$,$\hat{V_i}$,$\hat{D_i}$ are the sub-bands of the predicted image in the $i$-th downsampling, $L$ is the total level of downsampling.

We also add the spatial domain loss named $\mathcal{L}_{GT}$: let the ground-truth image be $Y$, the predicted image be $\hat{Y}$, and $\mathcal{L}_{GT}$ is the MSE between them:


\begin{equation}
    \mathcal{L}_{GT} = \mathbb{E}[\left\| Y - \hat{Y} \right\|^2]
\end{equation}

The total loss function of pre-trained CNN thus is:


\begin{equation}
    \mathcal{L}_{CNN} = \mathcal{L}_{GT} + \alpha \mathcal{L}_{FFT} +  \beta \mathcal{L}_{DWT}
\end{equation}

where $\alpha$ and $\beta$ are adjustable hyperparameters.

Furthermore, we design a simple CNN using residual-connection \cite{resnet} and pixel-shuffle \cite{pixel-shuffle}, named \textbf{SimpleSR}, for initial prediction (the specific structure is given in the supplementary material). Ablation studies on the proposed loss function and SimpleSR are given in the supplementary material.

\subsection{FD-guided Diffusion}

\begin{figure*}[t]
\begin{center}
\includegraphics[width=18 cm]{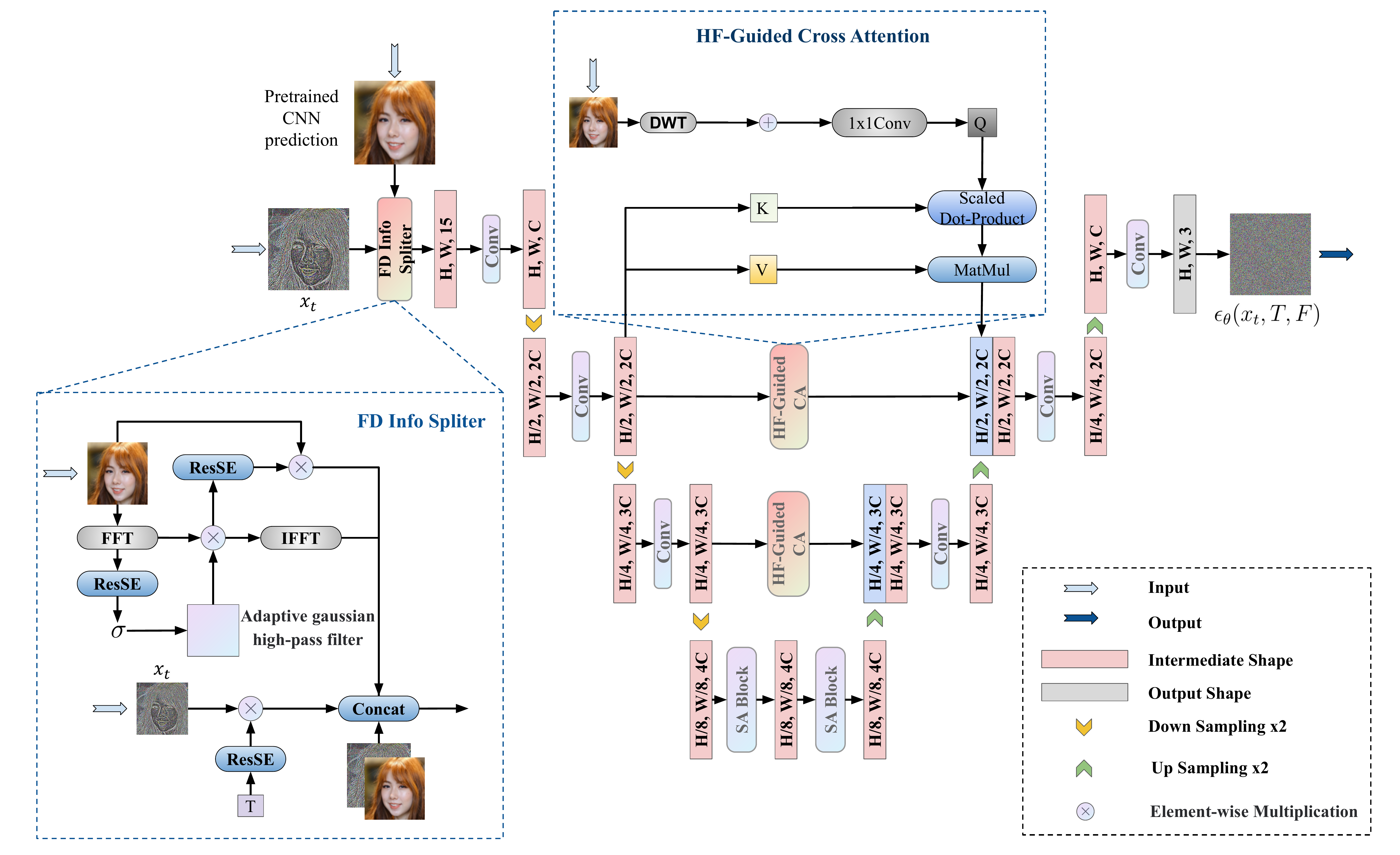}
\end{center}
\caption{An overview of the model architecture in proposed FD-guided diffusion. The pre-trained CNN prediction and the noisy image $x_t$ from step $t$ are fed into the FD-info-Splitter, and its output is then passed on to a U-net, which is equipped with HF-guided cross-attention.}
\label{fig:unet}
\end{figure*}

After obtaining the image $I$ predicted by the pre-trained CNN, we adapt a diffusion model to predict the residuals between $I$ and the ground truth, i.e., the high-frequency components of the ground-truth image. To this end, we propose a Frequency-Domain guided diffusion (FD-guided diffusion), as shown in Fig.\ref{fig:unet}. In contrast to SR3 \cite{sr3}, which simply concatenates the bilinear interpolated image with the noisy image $x_t$ at step $t$, we propose a Frequency-Domain Information Splitter module (FD-Info-Splitter): $I$ and $x_t$ is first fed into the FD-Info-Splitter, whose output is then fed into the U-net \cite{unet}. We follow the Imagen \cite{imagen}, where the self-attention layer is added. In addition, a Frequency-Domain guided Cross-Attention mechanism (FD-guild CA) is designed, which utilizes the high-frequency features obtained from DWT at each layer to generate more fine-grained detail features.

\subsection{FD Info Splitter}

For CNN's initial prediction, low-frequency components are mixed with high-frequency contents. As the diffusion model only needs to recover high-frequency details, the input low and high-frequency features have different statuses: the former mainly assist the generation of high-frequency components globally, while the latter is required to provide guidance for fine-grained details in each region. Therefore, we introduce Frequency-Domain Information Splitter (FD Info Splitter), which explicitly separates high-frequency and low-frequency information for better restoration. Additionally, it effectively mitigates noise for noisy images with large time steps, resulting in better noise prediction (The detailed structure of FD Info Splitter is shown in Fig.\ref{fig:unet}).

For the CNN predicted images $x_{cnn} \in \mathbb{R}^{H \times W \times C}$, we first perform 2D FFT along the spatial dimensions to obtain the frequency domain feature map $M$:


\begin{equation}
    M= FFT(x_{cnn}) \in \mathbb{C}^{H \times W \times C}
\end{equation}

where $FFT(\cdot )$ denotes the 2D FFT. We adapt the methods proposed by \cite{senet,resnet} and merged them into the ResSE module (Residual Squeeze-and-Excitation module), the details of which are shown in the supplementary material.

To implement adaptive high-pass filtering, a Gaussian high-pass filter is utilized whose Standard deviation is obtained from $M$ as follows:


\begin{equation}
    \sigma = min( |ResSE(M)| + \frac{l}{2} ,l )
\end{equation}

where $l = min(H,W)$. The operation for the acquired $ResSE(M)$ is for numerical stability. After obtaining $\sigma$, adaptive gaussian high-pass filter can be given directly as:

\begin{equation}
    H(u, v) = 1 - e^{{-D^2(u, v)}/{ (2 \sigma^ 2) }}
\end{equation}

where $D(u, v)$ is the distance from the point $(u, v)$ in the frequency domain to the center point. The gaussian high-pass filter are then preformed element-wise multiplication with $M$ to obtain the adaptive high-pass filtered feature map $M^{'}$:


\begin{equation}
    M^{'} = A_{hp} \otimes   M
\end{equation}

Finally, we reverse $M^{'}$ back to the spatial domain by adopting inverse FFT to obtain an feature map $x_{HF}$ rich in high-frequency components:


\begin{equation}
    x_{HF} = FFT^{-1}(M^{'})  \in \mathbb{R}^{H \times W \times C}
\end{equation}

where $FFT^{-1}(\cdot )$ denotes the Inverse 2D FFT. Meanwhile, we feed $M^{'}$ into a ResSE module to acquire the attention weights learned in the frequency domain and then perform element-wise multiplication with $x_{cnn}$ to obtain a feature map $x_{LF}$ containing abundant low-frequency information:


\begin{equation}
    x_{LF} = ResSE(M) \otimes  x_{cnn} 
\end{equation}

These two feature maps, dominated by high-frequency and low-frequency components, are concatenated in the channel dimension. By explicitly separating the input's mixed high-frequency and low-frequency components, the network can utilize both differently and more efficiently. 

For a noisy image $x_t$ at a large time step $t$, the noise components can be so large that it hinders network inference. Hence, an adaptive denoising is utilized on $x_t$ to obtain the partially denoised noisy image $x_t^{'}$:


\begin{equation}
    x_t^{'} = ResSE(T) \otimes  x_t 
\end{equation}

The three feature maps $x_{HF}$, $x_{LF}$, $x_t^{'}$, along with $x_{cnn}$ and $x_t$, are all concatenated in the channel dimension and fed into the U-net.


\subsection{HF-guided CA}


In the original U-net architecture, the encoder features are directly concatenated with the features obtained by the decoder \cite{unet}. This fusion facilitates the network to integrate the higher and lower-layer features effectively but lacks the ability to extract high-frequency features. To tackle this issue, we introduce a High-Frequency feature guided Cross-Attention mechanism (HF-guided CA) to recover fine-grained high-frequency details. The flow of the HF-guided CA is illustrated in Fig.\ref{fig:unet}.

We utilize the pre-trained CNN prediction by extracting the $\hat{H_i}$, $\hat{V_i}$, and $\hat{D_i}$ coefficients at the $i$-th level of the DWT. By adding these extracted coefficients with a linear projection, we obtain the feature map $Q$ with aggregated high-frequency information:

\begin{equation}
    Q=Conv_{1\times 1}(\hat{H_i}+\hat{V_i}+\hat{D_i})
\end{equation}

Then, different linear projections of the input feature map $M$ are constructed to obtain $K$ and $V$ in the cross-attention mechanism \cite{cross_attention} :


\begin{equation}
    K = Conv_{1\times 1}(M)
\end{equation}


\begin{equation}
    V = Conv_{1\times 1}(M)
\end{equation}

The output feature map $M^{'}$ can then be obtained from the formula:

\begin{equation}
    M^{'}=Softmax(\frac{QK^T}{\sqrt{d_k}})V
\end{equation}

where $d_k$ is the number of columns of matrix $Q$.






\section{Experiments}

\begin{figure*}[t]
\begin{center}
\includegraphics[width=16 cm]{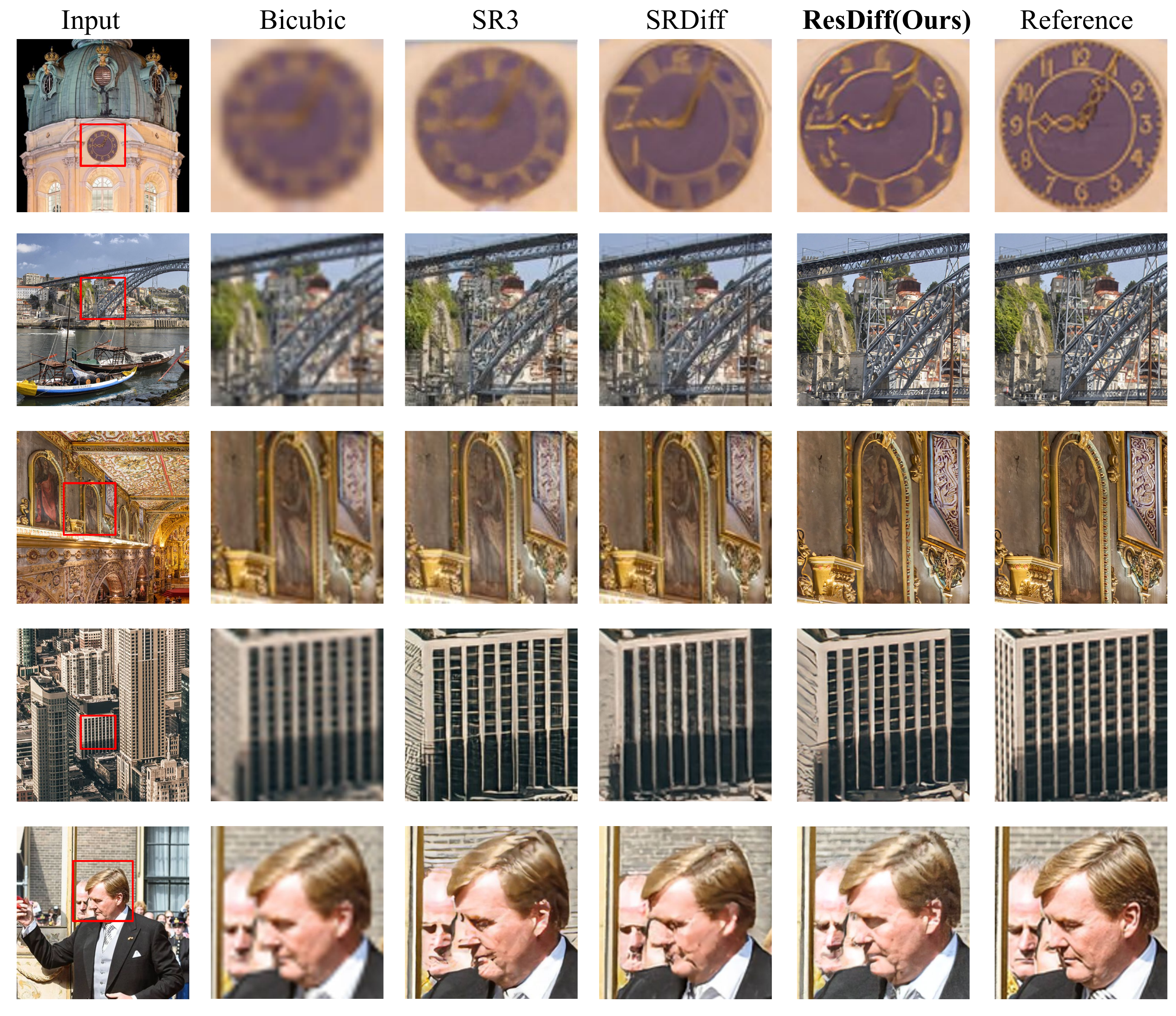}
\end{center}
\caption{DIV2k 4$\times$ results. Note that ResDiff provides richer details and more natural textures than other diffusion-based methods for the recovery of small objects (e.g., the clock in the first column) and difficult scenes (e.g., the bridge structure in the second column, the building in the fourth column).}
\label{fig:res_compare}
\end{figure*}





\subsection{Performance}

To evaluate the performance of our ResDiff model, we compared it with previous diffusion-based and GAN-based methods using four datasets: two face datasets (FFHQ \cite{ffhq}, and CelebA \cite{celeba}) and two general datasets (Div2k \cite{div2k}, and Urban100 \cite{urban100}). The selected evaluation metrics include two distortion-based metrics (PSNR and SSIM \cite{ssim}), as well as a perceptual-based metric (FID \cite{fid}). Our ResDiff is trained solely on the provided training data to guarantee a fair comparison. The supplementary material contains detailed information about the training process, hyperparameters, and other relevant details. Since several methods did not state their performance on some datasets we use, their values are marked as "-" in the table. More experiments with different types of datasets are presented in the supplementary material.

\paragraph{FFHQ and CelebA Results}

\begin{table}[htbp]
\small
\setlength{\tabcolsep}{1pt}
\centering
\begin{tabular}{lcccccc}
\toprule

& \multicolumn{3}{c}{$32 \to 128 $}
& \multicolumn{3}{c}{$256 \to 1024 $} \\

\cmidrule(lr){2-4} \cmidrule(lr){5-7}

& PSNR$\uparrow$
& SSIM$\uparrow$
& FID$\downarrow$
& PSNR$\uparrow$
& SSIM$\uparrow$
& FID$\downarrow$ \\

\midrule

Ground Truth
& $\infty$   & 1.000  & 0.00    
& $\infty$   & 1.000  & 0.00     \\

\cmidrule{1-7}


SRGAN
& 17.57    & 0.688     & 156.07     
& 21.49    & 0.515     & 60.67  \\

ESRGAN 
& 15.43    & 0.267     & 166.36  
& 19.84    & 0.353     & 72.73  \\
	
BRGM
& 24.16    & 0.70     &  -
&   -       &   --       &   -  \\

PULSE
& 15.74    & 0.37     &   -
&    -      &     -     & - \\

SRDiff
& {26.07}    & {0.794}     
& {72.36}
& {23.01}    & {0.656}   
& {56.17}\\

SR3
& {25.37}    & {0.778}     
& {75.29}
& {22.78}    & {0.647}   
& {60.12}\\

\cmidrule{1-7}

ResDiff
& \textbf{26.73}  & \textbf{0.818}  
& \textbf{70.54} 
& \textbf{23.15}   & \textbf{0.668}
& \textbf{53.23} \\

\bottomrule

\end{tabular}

\caption{Quantitative comparison on the FFHQ \cite{ffhq} dataset, where the bolded values represent the best value in each evaluation metric.}
\label{table:ffhq_res}

\setlength{\belowcaptionskip}{10pt}

\end{table}

\begin{table}[htbp]
\small
\setlength{\tabcolsep}{1pt}
\centering
\begin{tabular}{lcccccc}
\toprule

& \multicolumn{3}{c}{$20 \to 160 $}
& \multicolumn{3}{c}{$64 \to 256 $} \\

\cmidrule(lr){2-4} \cmidrule(lr){5-7}

& PSNR$\uparrow$
& SSIM$\uparrow$
& FID$\downarrow$
& PSNR$\uparrow$
& SSIM$\uparrow$
& FID$\downarrow$ \\

\midrule

Ground Truth
& $\infty$   & 1.000  & 0.00       
& $\infty$   & 1.000  & 0.00     \\

\cmidrule{1-7}


ESRGAN
& 23.24    & 0.66      &     -
&   -   &  -   & -  \\

PULSE
&     -          &  -         &   -
& 22.74         &  0.623        & 40.33 \\

SRFlow
& 25.28    & 0.72     &   -
&   -       &   -       &  - \\

SRDiff
& 25.32    & 0.73     &   80.98
& 26.84      &   0.792       & 39.16 \\

SR3
&24.89     & 0.728    &  83.11
&26.04     &  0.779   &   43.27\\

\cmidrule{1-7}

ResDiff
& \textbf{25.37}   & \textbf{0.734}   &  \textbf{78.52}
& \textbf{27.16}   & \textbf{0.797}   & \textbf{38.47} \\

\bottomrule

\end{tabular}

\caption{Quantitative comparison on the CelebA \cite{celeba} dataset, where the bolded values represent the best value in each evaluation metric.}

\label{table:celeba_res}

\setlength{\belowcaptionskip}{10pt}

\end{table}


The quantitative results at $32 \times 32 \to 128 \times 128$ ($4\times$) ,$256 \times 256 \to 1024 \times 1024$ ($4\times$) on FFHQ \cite{ffhq} and $20 \times 20 \to 160 \times 160$ ($8\times$), $64 \times 64 \to 256 \times 256$ ($4\times$) on CelebA \cite{celeba} are shown in table \ref{table:ffhq_res},\ref{table:celeba_res}. Our ResDiff demonstrates superior performance compared to all diffusion-based methods, as evidenced by the metrics presented in the table, and has about 50\% reduction in Perceptual metrics (FID) than the GAN-based model.


\paragraph{DIV2K and Urban100 Results}

\begin{table}[ht]
\small
\setlength{\tabcolsep}{1pt}
\centering
\begin{tabular}{lcccccc}
\toprule

& \multicolumn{3}{c}{ DIV2K $4 \times$}
& \multicolumn{3}{c}{ Urban100 $4 \times$} \\

\cmidrule(lr){2-4} \cmidrule(lr){5-7}

& PSNR$\uparrow$
& SSIM$\uparrow$
& FID$\downarrow$
& PSNR$\uparrow$
& SSIM$\uparrow$
& FID$\downarrow$ \\

\midrule

Ground Truth
& $\infty$   & 1.000  & 0.00      
& $\infty$   & 1.000  & 0.00     \\

\cmidrule{1-7}

SRDiff
& 26.87    & 0.69     &   110.32
&    26.49      &   0.79      &  51.37 \\

SR3
& 26.17    & 0.65    &  111.45
&  25.18     &  0.62   &   61.14 \\

\cmidrule{1-7}

ResDiff
&\textbf{27.94}   & \textbf{0.72}   &  \textbf{106.71}
& \textbf{27.43}      & \textbf{0.82}   & \textbf{42.35} \\

\bottomrule

\end{tabular}

\caption{Quantitative comparison on the DIV2K \cite{div2k} and Urban100 \cite{urban100} dataset, where the bolded values represent the best value in each evaluation metric.}

\label{table:general_res}

\setlength{\belowcaptionskip}{10pt}

\end{table}

The quantitative results at $40 \times 40 \to 160 \times 160$ ($4\times$) on DIV2K \cite{div2k} and $40 \times 40 \to 160 \times 160$ ($4\times$) on Urban100 \cite{urban100} are shown in table \ref{table:general_res}. Note that ResDiff's distortion-based metric values can significantly outperform other diffusion-based methods on these general datasets whose restoration is more difficult. Fig.\ref{fig:res_compare} presents partial results of ResDiff and other diffusion-based methods.





\subsection{Ablation Study}

In this section, we perform an ablation study on FFHQ ($4\times$) to investigate the effectiveness of each component in ResDiff, including the influence of different CNNs, and the usefulness of the proposed FD Info Splitter/HF-guided CA. The results are shown in Table \ref{table:compare_cnn}. Note that utilizing the residual structure, even with a simple bilinear interpolation for the initial prediction, can significantly improve the performance. In terms of CNN selection, our proposed SimpleSR also outperforms SRCNN \cite{srcnn}. Moreover, the addition of FD Info Splitter and HF-guided CA both have an improvement in the results. More detailed ablation studies are given in the supplementary material.

\section{Conclusion and Future Work}

In this paper, we propose ResDiff, a residual structure-based diffusion model. In contrast to the previous works, which only adapt LR images to generate HR images, ResDiff utilizes the feature-richer CNN prediction for guidance. Meanwhile, we introduce a frequency-domain-based loss function to the CNN and design a frequency-domain guided diffusion to facilitate the diffusion model in generating low-frequency information. Comprehensive experiments on different datasets demonstrate that the proposed ResDiff accelerates the training convergence speed and provides superior image generation quality.

Our ResDiff can also be adapted for other image restoration tasks, such as image blind super-resolution, deblurring, and inpainting. Although ResDiff can accelerate convergence, operations such as DWT are still time-consuming and call for optimization in future work. In addition, it can be seen from the supplementary material that the color will appear a large discrepancy when the model is under-trained, which may be caused by a lack of color features in the guided high-frequency information. Utilizing a global color feature may well address this issue in future work. Moreover, our ResDiff does not outperform current State-Of-The-Art(SOTA) SISR methods \cite{sota1,sota2}. This is attributed to the disparity between model parameters. Due to equipment limitations, adopting a larger U-net model in ResDiff is left to future work. In addition, if a pre-trained SOTA model is applied to replace the CNN in ResDiff, it may be possible to establish a new SOTA. Finally, ResDiff may consider incorporating more DPM techniques \cite{latent-diff,guilded-diff,free-guilded-diff} and superior network architectures \cite{better-network,better-network2} in the future.

\begin{table}[tb]
\small
\setlength{\tabcolsep}{1pt}
\centering
\tabcolsep=0.08cm
\begin{tabular}{lcp{1.8cm}<{\centering}cccc}
\toprule

& \multicolumn{3}{c}{Model Components}
& \multicolumn{3}{c}{Metrics} \\

\cmidrule(lr){2-4} \cmidrule(lr){5-7}

& \multirow{2}{*}{CNN}
& FD Info 
& HF-guided 
& \multirow{2}{*}{PSNR$\uparrow$}
& \multirow{2}{*}{SSIM$\uparrow$}
& \multirow{2}{*}{FID$\downarrow$} \\

& ~
& Splitter 
& CA
& ~
& ~
& ~ \\

\midrule

&  SimpleSR  & $\checkmark$      &   $\checkmark$
& 26.73     & 0.818     & 70.54 \\

\cmidrule{1-7}

& N/A  & $\checkmark$  & $\checkmark$     
& 25.49    & 0.781  & 74.18 \\

& Bilinear     & $\checkmark$      &   $\checkmark$ 
& 25.99    &  0.792   & 74.29   \\

& SRCNN        & $\checkmark$      &   $\checkmark$ 
&  26.14   &  0.809   &  72.17  \\

& SimpleSR  & \multirow{2}{*}{$\checkmark$}      &\multirow{2}{*}{$\checkmark$} 
&  \multirow{2}{*}{26.47}   &  \multirow{2}{*}{0.812}    &  \multirow{2}{*}{71.58}  \\

& (only $\mathcal{L}_{GT}$)   & ~      &   ~
&  ~   & ~    &  ~  \\

\cmidrule{1-7}

& SimpleSR       &        &    
& 25.41    & 0.788    & 77.21  \\

& SimpleSR   &      &   $\checkmark$ 
& 26.09    &  0.796   &  72.42     \\

& SimpleSR   & $\checkmark$      &   
&  25.97   &  0.793   &  73.17  \\

\bottomrule

\end{tabular}

\caption{Ablation study over different model components on the ffhq \cite{ffhq} test sets (The model components we use are placed in the first row). N/A denotes no residual structure used.}

\label{table:compare_cnn}

\setlength{\belowcaptionskip}{10pt}

\end{table}

\section{Acknowledgments}
We gratefully thank the creators of the dataset and the server support from Shandong University and Linyi University. This work was supported in part by the National Key Research and Development Program of China under Grant 2022YFB4500602, the Key Research and Development Program of Jiangsu Province under Grant BE2021093, Distinguished Young Scholar of Shandong Province under Grant ZR2023JQ025, Taishan Scholars Program  under Grant tsqn202211290, and Major Basic Research Projects of Shandong Province under Grant ZR2022ZD32.

\bibliography{aaai24}

\end{document}